\documentclass[11pt,a4paper]{article}
\usepackage[hyperref]{eacl2021}
\usepackage{times}
\usepackage{latexsym}

\usepackage{amsmath, amssymb}
\usepackage{breqn}
\usepackage{tabularx}
\usepackage{graphicx}
\usepackage{booktabs, makecell}
\usepackage{footnote}
\usepackage{adjustbox}
\usepackage{multirow}
\usepackage{algorithm}
\usepackage[noend]{algpseudocode}
\usepackage{threeparttable}
\usepackage{subcaption,siunitx,booktabs}
\usepackage{mathtools}
\usepackage{nidanfloat}

\usepackage{microtype}

\aclfinalcopy 
\def\aclpaperid{4} 

\title{Text Simplification by Tagging}

\author{Kostiantyn Omelianchuk\thanks{\hspace{4pt}Authors contributed equally to this work; names are given in alphabetical order.} \\ \And
  Vipul Raheja\footnotemark[1] \\
  Grammarly \\
  \texttt{firstname.lastname@grammarly.com} \\\And
  Oleksandr Skurzhanskyi\footnotemark[1]}

\date{}

\begin{document}
\maketitle
\begin{abstract}
Edit-based approaches have recently shown promising results on multiple monolingual sequence transduction tasks. In contrast to conventional sequence-to-sequence (Seq2Seq) models, which learn to generate text from scratch as they are trained on parallel corpora, these methods have proven to be much more effective since they are able to learn to make fast and accurate transformations while leveraging powerful pre-trained language models. Inspired by these ideas, we present TST, a simple and efficient \textbf{T}ext \textbf{S}implification system based on sequence \textbf{T}agging, leveraging pre-trained Transformer-based encoders. Our system makes simplistic data augmentations and tweaks in training and inference on a pre-existing system, which makes it less reliant on large amounts of parallel training data, provides more control over the outputs and enables faster inference speeds. Our best model achieves near state-of-the-art performance on benchmark test datasets for the task. Since it is fully non-autoregressive, it achieves faster inference speeds by over 11 times than the current state-of-the-art text simplification system. 
\end{abstract}

\section{Introduction}
Text Simplification is the task of rewriting text into a form that is easier to read and understand while preserving its underlying meaning and information. It has been shown to be valuable in providing assistance in terms of readability and understandability to children \cite{Belder2010TextSF, kajiwara-etal-2013-selecting}, people with language disabilities like aphasia \cite{carroll1998practical, carroll-etal-1999-simplifying, 10.1145/1168987.1169027}, dyslexia \cite{10.1145/2461121.2461126, 10.1007/978-3-642-37256-8_41}, or autism \cite{evans-etal-2014-evaluation}; non-native English speakers \cite{Petersen07textsimplification, paetzold-2015-reliable, paetzold-specia-2016-understanding, 10.5555/3016387.3016433, pellow-eskenazi-2014-open}, and people with low literacy skills or reading ages \cite{10.1007/11671299_59, 10.1145/1456536.1456540, Gasperin2009NaturalLP, 10.1145/1621995.1622002}.
Moreover, it has also been successfully leveraged as a pre-processing step to improve the performance of various NLP tasks such as parsing \cite{chandrasekar-etal-1996-motivations}, summarization \cite{10.1007/978-3-540-30468-5_47, Silveira12enhancingmulti-document}, semantic role labeling \cite{vickrey-koller-2008-sentence, 10.5555/3171837.3172021} and machine translation \cite{10.1007/3-540-49478-2_40, stajner-popovic-2016-text, HASLER2017221}.

Evolving from the approaches ranging from building hand-crafted rules \cite{chandrasekar-etal-1996-motivations, Siddharthan2006} to syntactic and lexical simplification via synonyms and paraphrases \cite{oro58886, kaji-etal-2002-verb, horn-etal-2014-learning, glavas-stajner-2015-simplifying}, the task has gained popularity as a monolingual Machine Translation (MT) problem, where the system learns to ``translate" a given complex sentence to its simplified form. Initially, Statistical phrase-based (SMT) and Syntactic-based Machine Translation (SBMT) techniques  \cite{zhu-etal-2010-monolingual, 10.1007/978-3-642-12320-7_5, coster-kauchak-2011-learning, wubben-etal-2012-sentence, narayan-gardent-2014-hybrid, stajner-etal-2015-deeper, xu-etal-2016-optimizing} were successfully applied as a way to learn simplification rewrites implicitly from complex-simple sentence pairs, often in combination with hand-crafted rules or features. More recently, several Neural Machine Translation-based (NMT) systems have been developed with promising results \cite{NIPS2014_5346, cho-etal-2014-learning, DBLP:journals/corr/BahdanauCB14}, and their successful application to text simplification, either in combination with SMT or other data-driven approaches \cite{Zhang2017ACS, zhao-etal-2018-integrating}; or strictly as neural models \cite{AAAI1611944, nisioi-etal-2017-exploring, zhang-lapata-2017-sentence, stajner-nisioi-2018-detailed, guo-etal-2018-dynamic, vu-etal-2018-sentence, 10.1007/978-3-030-04221-9_48, kriz-etal-2019-complexity, surya-etal-2019-unsupervised, DBLP:conf/aaai/ZhaoCCY20}, has emerged as the state-of-the-art. 

Human editors perform several rewriting transformations in order to simplify a sentence, such as lexical paraphrasing, changing the syntactic structure, or removing superfluous information from the sentence \cite{Petersen07textsimplification, Alusio2008ACA, mallinson2020felix}. Therefore, even though NMT-based sequence-to-sequence (Seq2Seq) approaches offer a generic framework for modeling almost any kind of sequence transduction, target texts in these approaches are typically generated from scratch - a process which can be unnecessary for monolingual editing tasks such as text simplification, owing to these aforementioned transformations. Moreover, these approaches have a few shortcomings that make them inconvenient for real-world deployment. First, they give limited insight into the simplification operations and provide little control or adaptability to different aspects of simplification (e.g., lexical vs. syntactical simplification). This inhibits interpretability and explainability, which is crucial for real-world settings. Second, they are not sample-efficient and require a large number of complex-simple aligned sentence pairs for training, which requires considerable human effort to obtain. Third, these models typically employ an autoregressive decoder, i.e., output texts are generated in a sequential, non-parallel fashion, and hence, are generally characterized by slow inference speeds.

Based on the aforementioned observations and issues, text-editing approaches have recently re-gained significant interest \cite{NIPS2019_9297, dong-etal-2019-editnts, awasthi-etal-2019-parallel,  malmi-etal-2019-encode, omelianchuk-etal-2020-gector, mallinson2020felix}. Typically, the set of edit operations in such tasks is fixed and predefined ahead of time, which on one hand limits the flexibility of the model to reconstruct arbitrary output texts from their inputs, but on the other, leads to higher sample-efficiency as the limited set of allowed operations significantly reduces the search space \cite{mallinson2020felix}. This pattern is especially true for monolingual settings where input and output texts have relatively high degrees of overlap. In such cases, a natural approach is to cast the task of conditional text generation into a text-editing task, where the model learns to reconstruct target texts by applying a set of edit operations to the inputs. We leverage this insight in our work, and simplify the task from sequence generation or editing, going a step further, to formulate it as a  sequence tagging task. In addition to being sample efficient, thanks to the separation of various edit operations in the form of tags, the system has better interpretability and explainability. Finally, since for sequence tagging we don’t need to predict tokens one-by-one as in autoregressive decoders, the inference is naturally parallelizable and therefore runs many times faster.

Following from the success of the aforementioned monolingual edit-tag based systems, we propose to leverage the current state-of-the-art model for Grammatical Error Correction by \citet{omelianchuk-etal-2020-gector} (GECToR) and adapt it to the task of Text Simplification. 
In summary, we make the following contributions:
\begin{itemize}
    \itemsep0pt
    \item We develop a Text Simplification system by adapting the GECToR model to Text Simplification, leveraging Transformer-based encoders trained on large amounts of human-annotated and synthetic data.\footnote{Available at \url{https://github.com/grammarly/gector\#text-simplification}} Empirical results demonstrate that our system achieves near state-of-the-art performance on benchmark test datasets in terms of readability and simplification metrics.
    \item We propose crucial data augmentations and tweaks in training and inference and show their significant impact on the task: enabling the model to learn to edit the sentences more effectively, rather than relying heavily on copying the source sentences, leading to a higher quality of simplifications.
    \item Since our model is a non-autoregressive sequence tagging model, it achieves over 11 times speedup in inference time, compared to the state-of-the-art for Text Simplification.
\end{itemize}

\section{Related Work}
Recent text editing works have shown promising results of reformulating multiple monolingual sequence transduction tasks into sequence tagging tasks compared to the conventional Seq2Seq sequence generation formulation. This observation is especially true for tasks where input and output sequences have a large overlap. 
Generally, these works try to simplify monolingual sequence transduction by explicitly modeling edit operations such as \textsc{keep}, \textsc{add}/\textsc{insert} and \textsc{delete}. \citet{alva-manchego-etal-2017-learning} proposed the first such formulation, employing a BiLSTM to predict edit labels sequentially. Our model for sentence simplification does not rely on external simplification rules nor alignment tools. \citet{ribeiro-etal-2018-local} proposed an approach applied only to character deletion and insertion and was based on simple patterns. LaserTagger \cite{malmi-etal-2019-encode} combines a BERT encoder with an autoregressive Transformer decoder to similarly predict the aforementioned three main edit operations for several text editing tasks. In contrast, in our system, the decoder is a softmax layer. Similarly, EditNTS \cite{dong-etal-2019-editnts} and PIE \cite{awasthi-etal-2019-parallel} predict edit labels, developed specifically for text simplification and GEC, respectively. While EditNTS employs an autoregressive encoder-decoder based neural programmer-interpreter model, PIE differs from our work because of our custom edit transformations and incorporation of a pre-trained Transformer encoder for sequence tagging. Levenshtein Transformer \cite{NIPS2019_9297}, an autoregressive model that performs text editing by executing a sequence of deletion and insertion actions, is another recent work along similar lines. More recently, \citet{mallinson2020felix} proposed Felix - a text-editing-based system for multiple generation tasks, splitting the text-editing task into two sub-tasks: tagging and insertion. Their tagging model employs a Pointer mechanism, while the insertion model is based on a Masked Language Model. 

\section{System Description}

\begin{figure}
    \centering
    \includegraphics[width=\columnwidth]{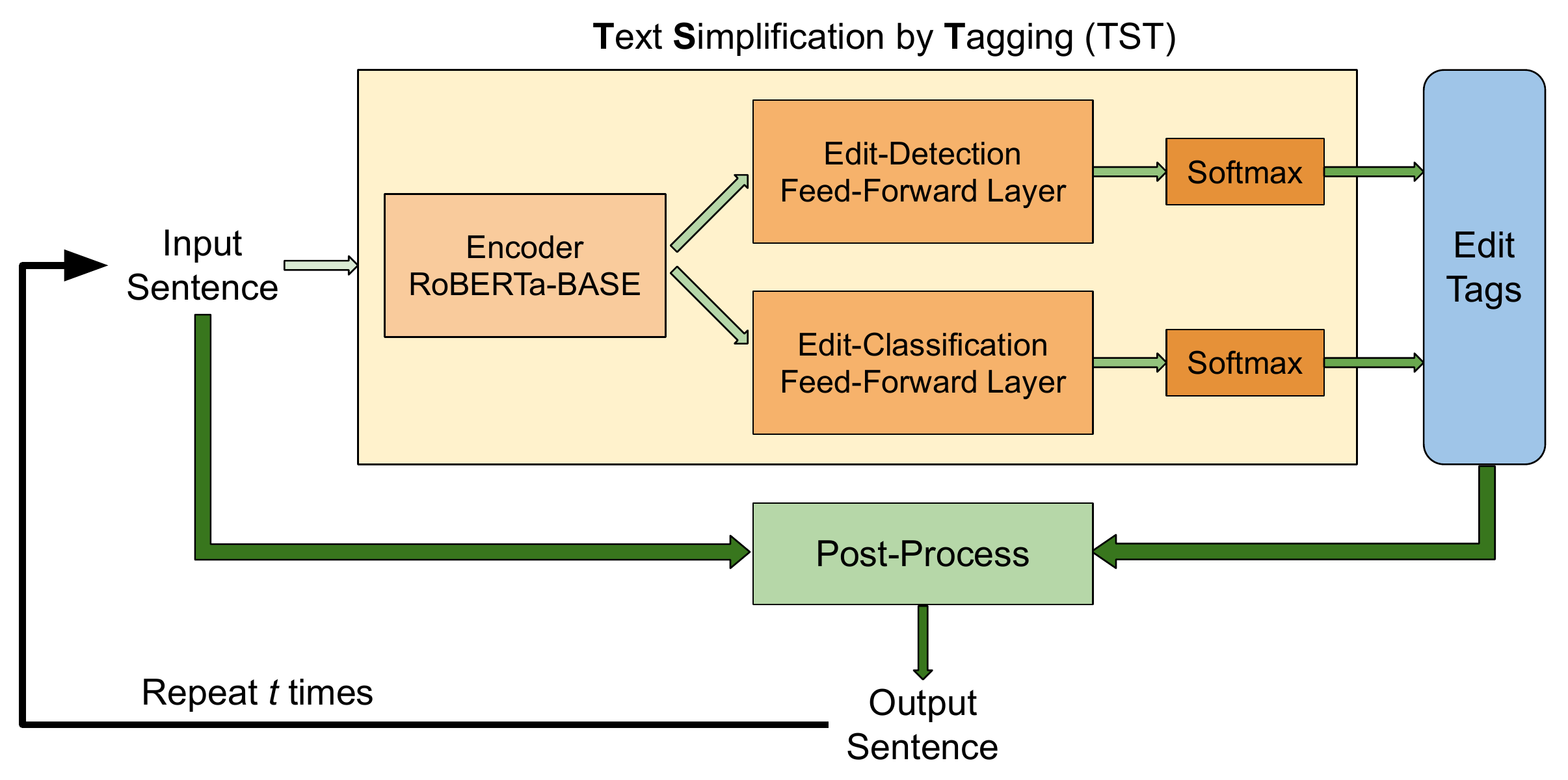}
    \caption{\textbf{T}ext \textbf{S}implification by \textbf{T}agging (TST): A given sentence undergoes multiple iterations of tag-and-edit transformations, where, in each iteration, it is tagged using custom token-level edit-tags, and the sequence of tags is converted back to text by applying those edits, iteratively making simplifying edits.}
    \label{fig:framework}
\end{figure}

Following recent works such as \citet{malmi-etal-2019-encode, awasthi-etal-2019-parallel, omelianchuk-etal-2020-gector}, who leveraged similar frameworks for different text editing problems such as GEC, Sentence Fusion, and Abstractive Summarization, we formulate the task of Text Simplification as a tagging problem.

Specifically, our system is based on GECToR \cite{omelianchuk-etal-2020-gector}, an iterative sequence-tagging system that works by predicting token-level edit operations, originally developed for Grammatical Error Correction (GEC). 
We adapt the GECToR framework for the task of Text Simplification, with minimal modifications to the original architecture. Our system consists of three main parts: (a) defining the custom transformations (token-level edit-tags), (b) performing iterative sequence tagging to convert target sequences to tag sequences, (c) fine-tuning of pre-trained Transformers to predict the tag sequences. Each of these components are described below.

\subsection{Edit Transformations}
\label{section:transformations}
In order to formulate the task as a tagging problem, building on the aforementioned edit-tagging-based approaches, we use custom token-level edit operations (also referred to as \textbf{edit-tags} or \textbf{transformations}) to perform text simplification. Formally, given a sentence $x$: $[x_1, x_2, \dots , x_N]$, and its simplified form $y$: $[y_1, y_2, \dots , y_M]$ as the target sentence, we aim to predict an edit tag $t_i \in \tau$ ($\tau$ denoting the edit-tag vocabulary) for each token $x_i$ in $x$, generating a sequence of edit-tags of the same length $N$ as the input sequence $x$, such that $t_i(x_i)$: applying the edit operation represented by the edit-tag $t_i$ to the input token $x_i$ at each position $i$, reconstructs the target sequence $y$, even though $M \leq N$.

We reuse the edit transformations in GECToR, which were developed for GEC. We chose to do so because we found a significantly high overlap of 92.64\% in the tag distributions between the GEC and Text Simplification domains. This was done by building the edit-tag vocabularies independently on both (GEC and Text Simplification) datasets and comparing the tag distributions represented by the two vocabularies. This was not surprising since these edit-tags have been obtained from huge amounts of synthetic GEC data, they are expected to have good coverage with many standard monolingual text editing problems. Additionally, using the same edit-tags is a necessary pre-requisite to leverage GEC initialization in the model (Section \ref{ssec:pre-training}), which we later show to be quite impactful for our text simplification system (Section \ref{ssec:ablation_gec}). Consequently, the edit space $\tau$ is of size 5000, out of which 4971 are basic edit-tags and 29 are token-independent GEC-specific edit-tags (such as \texttt{\$TRANSFORM_VERB_VB_VBZ}, which converts a verb in its base form to its third person singular present tense form). Further, the aforementioned 4971 basic edit-tags are made up of token-independent \textsc{keep} and \textsc{delete} tags (which simply keep or delete the given word(s) on which they are applied), 1167 token-dependent \textsc{append} tags (such as \texttt{\$APPEND_just}, which appends the word ``just" to the given word) and 3802 token-dependent \textsc{replace} tags (such as \texttt{\$REPLACE_really}, which replaces the given word with the word ``really").

\subsection{Iterative Sequence Tagging}
\label{ssec:iterative_tagging}
As described in Section \ref{section:transformations}, 
we predict the edit-tags $t_i$ for each input token $x_i$ in the source sequence $x$. These predicted tag-encoded transformations are then applied to the source sentence to get the simplified sentence. Since some simplification operations in a sentence may depend on others, applying the sequence tagger only once may not be enough to fully generate the simplified form of a given sentence. Accordingly, we use the iterative correction approach from \citet{awasthi-etal-2019-parallel} and \citet{omelianchuk-etal-2020-gector}, and use the sequence tagger to tag the now modified sequence, and apply the corresponding transformations on the new edit-tags, which changes the sentence further. We repeat this process for a fixed number of iterations, which can be adjusted to trade off qualitative performance for improved speed. In our framework, we experimented between 1-5 iterations.


\subsection{Tagging Model}
\label{ssec:tagging_model}
We use the GECToR sequence tagging model with a pre-trained RoBERTa$_{\textsc{BASE}}$ Transformer \cite{DBLP:journals/corr/abs-1907-11692} as the encoder, stacked with two concurrent feed-forward layers, followed by corresponding Softmax layers. Owing to our choice of encoder, we use Byte-Pair Encoding (BPE) \cite{sennrich-etal-2016-neural} as our tokenization technique. 

As shown in Fig. \ref{fig:framework}, these feed-forward layers are responsible for detecting and classifying edits, respectively. For every position in the input sequence, the edit-detection layer predicts the probability an edit exists, whereas the edit-classification layer predicts the type of edit-tag. 
The edit-tag sequence generated as the output of the edit-classification layer is gated by the output of the edit-detection layer. 
i.e. if the output of the edit-detection layer 
is below the \textit{minimum edit probability} threshold (described in Section \ref{ssec:inference_tweaks_description}) at any position in the predicted sequence, we do not make any edits. 



\section{Experimental Setup}
\label{sec:experimental_setup}

\subsection{Data Sources}
\label{section:data_sources}
We use \textbf{WikiSmall} and \textbf{WikiLarge}, two benchmark datasets for the text simplification task\footnotemark, for our experiments. These datasets were constructed from automatically-aligned complex-simple sentence pairs from English Wikipedia (EW) and Simple English Wikipedia (SEW). WikiSmall \cite{zhu-etal-2010-monolingual} contains one reference simplification per sentence. We use the standardized split of this dataset released by \citet{zhang-lapata-2017-sentence}, with 88\textit{k} instances for training, 205 for validation and the same original 100 instances for testing. WikiLarge is a larger set of similarly automatically-aligned complex-simple sentence pairs, compiled from previous extractions of EW-SEW and WikiSmall \cite{zhu-etal-2010-monolingual, woodsend-lapata-2011-learning, kauchak-2013-improving}. Similar to WikiSmall, we use the training set for this dataset provided by \citet{zhang-lapata-2017-sentence} consisting of 296\textit{k} sentence pairs. For simplicity, we refer to this training data (WikiSmall + WikiLarge) as \textbf{WikiAll}.

For validation and test sets, we use the Turkcorpus \cite{xu-etal-2016-optimizing} and ASSET \cite{alva-manchego-etal-2020-asset} datasets, which were both created from WikiLarge using the same 2000 validation and 359 test source sentences, where each complex sentence consists of multiple crowd-sourced reference simplifications. Specifically, Turkcorpus contains 8 reference simplifications, and ASSET contains 10 references per source sentence. Table \ref{tab:data-stats} provides other statistics on these datasets.
\footnotetext{Another widely-used dataset for the task, the Newsela Corpus \cite{xu-etal-2015-problems}, could not be used due to its extremely rigid legal and licensing requirements.}

\begin{table}[t!]
\begin{center}
\small
\resizebox{\columnwidth}{!}{%
\begin{tabular}{*{5}{l}}
\toprule
\textbf{Split} & \multicolumn{2}{c}{\textbf{Dataset}} & \textbf{Sentences} & \textbf{Tokens} \\
\toprule
\multirow{4}{*}{Train} 
    & \multirow{2}{*}{WikiAll} 
        & WikiSmall & 88\textit{k}& 3.9M\\
    &   & WikiLarge & 296\textit{k} & 11.7M \\ \cmidrule{2-5}
    & \multirow{2}{*}{WikiBT}
    & WikiLarge (En-De) & 293\textit{k} & 11.2M\\
    &   & WikiLarge (En-Fr) & 293\textit{k} & 11.5M\\
\midrule
\multirow{3}{*}{Dev} 
    & WikiSmall & & 205 & 9.5\textit{k}\\
    \cmidrule{2-5}
    & \multirow{2}{*}{WikiLarge}    
        & TurkCorpus & 2000 & 75\textit{k}\\
        & & ASSET & 2000 & 72\textit{k}\\
\midrule
\multirow{3}{*}{Test} 
    & WikiSmall & & 100 & 5\textit{k}\\
    \cmidrule{2-5}
    & \multirow{2}{*}{WikiLarge}
        & TurkCorpus & 359 & 15.8\textit{k}\\
        & & ASSET & 359 & 14.1\textit{k}\\
\bottomrule
\end{tabular}
}
\caption{Dataset splits and sizes.}
\label{tab:data-stats}
\end{center}
\end{table}

\subsection{Data pre-processing}
\label{ssec:data_preproc}
WikiAll data contains special tokens to represent parentheses (symbolized by \texttt{-LRB-} and \texttt{-RRB-}) from prior tokenizations. We heuristically decide to remove these tokens (and any tokens between them) from both source and target sentences. Doing this led to consistent improvements in all our experiments, described further in Section \ref{ssec:ablation_brackets}.
Additionally, for tokenization, we use the HuggingFace Tokenizers\footnote{\url{https://github.com/huggingface/tokenizers}} Python library to tokenize the whole sentence (as opposed to the approach in GECToR which tokenized each word in the sentence separately). This change led faster and more accurate tokenization as the one originally used in RoBERTa. 


    
    



\subsection{Pre-Training}
\label{ssec:pre-training}
For our experiments, we use two versions of the tagging model described in Section \ref{ssec:tagging_model}. The first version is a pre-trained RoBERTa$_{\textsc{BASE}}$ encoder with randomly initialized feed-forward layers. We refer to this model as \textbf{\textsc{TST-Base}} (\textbf{T}ext \textbf{S}implification by \textbf{T}agging - Baseline). The second version of the model is a \textsc{TST-Base} model fine-tuned on the  Grammatical Error Correction (GEC) task: henceforth denoted as \textbf{\textsc{TST-GEC}}.\footnote{We refer the reader to \citet{omelianchuk-etal-2020-gector} for details on training the model for GEC.}

\subsection{Data Augmentation}
We hypothesize that our text simplification models can benefit from an increase of the training data, and experimentally confirm this by training and evaluating our models with additional training data. We generate synthetic training data from the source sentences of WikiAll. We used two approaches to do so: back-translation and ensemble distillation, described below.

\subsubsection{Back-Translation}
We use the Transformer-based NMT models trained by \citet{TiedemannThottingal:EAMT2020} to generate the back-translated versions of the target side of the parallel WikiAll data. These models were trained on OPUS data\footnote{\url{http://opus.nlpl.eu/}} using Marian-NMT\footnote{\small{ \url{https://marian-nmt.github.io/}}} and released as part of the HuggingFace Transformers Library \cite{Wolf2019HuggingFacesTS}. All models are Transformer encoder-decoders with 6 layers in each component. Specifically, we used the bilingual \textsc{en-fr}, \textsc{fr-en}, \textsc{en-de} and \textsc{de-en} models\footnote{\small{\url{https://huggingface.co/Helsinki-NLP/opus-mt-<L1>-<L2>}}} to translate WikiAll data from (a) English to French, and back to English, and (b) English to German and back to English. Doing so tripled the amount of WikiAll data available for training (Table \ref{tab:data-stats}). The backtranslated WikiAll data is henceforth collectively referred to as \textbf{WikiBT}.

\subsubsection{Ensemble Distillation}
We leverage knowledge distillation on ensemble teacher models \cite{Freitag2017EnsembleDF} to augment our training data. We first create an ensemble teacher model by training models on WikiAll and WikiBT data. Specifically, for building the teacher ensemble, we first train the following constituent TST models:
\begin{enumerate}
    \item \textsc{TST}: Trained on WikiAll
    \item \textsc{TST-GEC}: Trained on WikiAll
    \item \textsc{TST}: Trained on WikiAll + WikiBT
\end{enumerate}

The predictions of the ensemble are computed by taking the \texttt{argmax} of the averaged class-wise probabilities of the constituent models at every token. 
We get the predictions from this ensemble consisting of the aforementioned three constituent models on WikiAll data. In this way, we produce new references for the training data which can be used by our final model (referred to as the student network) to simulate the teacher network ensemble. 
We then combine this ensemble-generated training data (hereby referred to as \textbf{WikiEns}) together with the original WikiAll data, doubling the amount of training data.
Our final model (the student network, denoted henceforth as \textbf{\textsc{TST-Final}}) is then trained on this combined WikiEns + WikiAll dataset. 
It is worth noting that the student and the constituent teacher models have exactly the same architecture. 

\subsection{Training}
We train our models with AllenNLP and Transformers.  
Our baseline (\textsc{TST-Base}) mostly follows the settings in \citet{omelianchuk-etal-2020-gector}. 
We train the model for 50 epochs, where we freeze the encoder weights during the first two epochs of training. We use Adam optimizer \cite{DBLP:journals/corr/KingmaB14}, where the learning rate starts from 1e-5 and reduces by a factor of 0.1 when the validation loss has stopped improving for 10 epochs. We perform early stopping after 3 epochs, based on the performance on the validation set. Other training hyper-parameters are listed in Appendix \ref{sec:appendix1}.

\subsection{Inference Tweaks}
\label{ssec:inference_tweaks_description}
One of the advantages of edit-tag-based approaches is that they provide greater control over the system output. Building on \citet{omelianchuk-etal-2020-gector}, we use \textit{confidence biases} and \textit{minimum edit probability} as additional inference hyper-parameters that we tune to push the model to perform more precise edits. 

Specifically, we add confidence biases to the probabilities of \textsc{KEEP} and \textsc{DELETE} edit-tags: responsible for not changing the source token and deleting the source token, respectively. We create these additional hyper-parameters for just these edit-tags because they are the most frequently used edit-tags for the Text Simplification task. Moreover, since they are token-independent, it provides the framework with additional robustness on the task without introducing too many additional hyper-parameters.
In this way, we were able to drive the model to keep/delete more tokens if the corresponding confidence bias was positive and to keep/delete fewer tokens if it was negative. We also add a sentence-level minimum edit probability threshold ($\epsilon$) for the output of the edit detection layer. This hyper-parameter enabled the model to predict only the most confident edits. Thus, we were able to increase precision by trading off the recall and achieve better performance.


These hyper-parameters were tuned using a combination of random search and Bayesian search \cite{bayesian-opt-nogueira} on the respective validation sets. Section \ref{sec:ablation_studies} further describes the impact of the aforementioned tweaks on the system. Final values of these hyper-parameters are listed in Appendix \ref{sec:appendix1}.


\subsection{Evaluation Metrics}
We report the results using two widely used metrics in Text Simplification literature: 
\textbf{FKGL} \cite{kincaid1975derivation}, and \textbf{SARI} \cite{xu2016optimizing}. Prior work has also used \textbf{BLEU} \cite{papineni-etal-2002-bleu} as a metric, but recent work has found that it is not a suitable metric for evaluating text simplification, because it was found to be negatively correlated with simplicity, 
essentially penalizing simpler sentences \cite{sulem-etal-2018-bleu}.

FKGL (\textbf{F}lesch-\textbf{K}incaid \textbf{G}rade \textbf{L}evel) is used to measure the readability of the generated sentence, where a lower score indicates simpler output. FKGL doesn't use source sentences or references for computing the score. It is a linear combination of the number of words per sentence (system output) and the number of syllables per word. On the other hand, SARI (\textbf{S}ystem output \textbf{A}gainst \textbf{R}eferences and against the \textbf{I}nput sentence) evaluates the quality of the output by comparing the generated sentence to a set of reference sentences in terms of correctly inserted, kept and deleted n-grams ($n\in{1, 2, 3, 4}$). 
We report the overall SARI metric, and scores on the three rewrite operations used in SARI: the F1-scores of add (\textsc{add}), delete (\textsc{delete}) and keep (\textsc{keep}) operations. 
FKGL and SARI are both measured at corpus-level. 
We computed all the evaluation metrics using the EASSE\footnote{\url{https://github.com/feralvam/easse}} Python package \cite{alva2019easse}. 

\section{Results}

\begin{table}[t!]
\centering

\begin{subtable}{.48\textwidth}
    \centering
    \resizebox{\textwidth}{!}{%
    \begin{threeparttable}[t]
    \begin{tabular}{l|c|ccc|c}
        \toprule
            & SARI$\uparrow$ & ADD$\uparrow$ & DELETE$\uparrow$ & KEEP$\uparrow$ & FKGL$\downarrow$  \\
        \midrule
        \midrule
        \textbf{Recent Works} \\
        \midrule
        \midrule
        \citet{xu2016optimizing} & 39.96 &	5.96 &	41.42 &	72.52 & 7.29 \\
        \citet{nisioi-etal-2017-exploring}	& 	35.66	& 	2.99	& 	28.96	& 	\textbf{75.02}	& 	8.42  \\
        \citet{zhang-lapata-2017-sentence}	& 	37.27	&  -	& 	-	& 	-	& 	6.62 \\
        \citet{alva-manchego-etal-2017-learning}\tnote{$\ddagger$} & 37.08 & 2.94 & 43.20 & 65.10 & \textbf{5.35} \\
        \citet{vu-etal-2018-sentence} & 36.88 & - & - & - & - \\
        \citet{zhao2018integrating}	& 	40.42	& 	5.72	& 	42.23	& 	73.41	& 	7.79   \\
        \citet{guo-etal-2018-dynamic}	& 	37.45	& 	-	& 	-	& 	-	& 	7.41 \\
        \citet{qiang2018improving} &   37.21   &   -	& 	-	& 	-	& 	6.56 \\
        \citet{surya-etal-2019-unsupervised}	& 	34.96	& 	-	& 	-	& 	-	& 	-  \\
        \citet{dong-etal-2019-editnts} & 	38.22	& 	3.36	& 	39.15	& 	72.13	& 	7.3	\\
        \citet{zhao2020semi} & 	37.25	& 	2.87	& 	40.06 & 68.82	& 	- \\
        \citet{mallinson2020felix} & 	38.13	& 	3.55	& 	40.45	& 	70.39	& 	8.98 \\
        \citet{martin-etal-2020-controllable} & 41.38 & - & - & - & 7.29 \\
        \citet{martin2020multilingual} & \textbf{42.53$_{\pm 0.36}$} & - & - & - & 7.60$_{\pm 1.06}$ \\
        \midrule
        \midrule
        Reference Baseline & 40.02$_{\pm 0.72}$ & 6.21$_{\pm 0.60}$ & \textbf{70.15$_{\pm 1.35}$} & 43.69$_{\pm 1.46}$ & 8.77$_{\pm 0.19}$ \\        
        \midrule
        \midrule
        \textbf{Our System} \\
        \midrule
        \textsc{TST-Base} & 39.17$_{\pm 0.77}$ & 3.62$_{\pm 0.41}$ & 41.61$_{\pm 3.14}$ & 72.29$_{\pm 1.45}$ & 8.08$_{\pm 0.31}$ \\
        \textsc{TST-Final}  & 41.46$_{\pm 0.32}$ & \textbf{6.96$_{\pm 0.44}$} & 47.87$_{\pm 0.75}$ & 69.56$_{\pm 1.19}$ & 7.87$_{\pm 0.19}$ \\
        \bottomrule
    \end{tabular}
    \begin{tablenotes}
        \item[$\ddagger$] Quoted from the re-implementation by \citet{dong-etal-2019-editnts}.
    \end{tablenotes}
    \end{threeparttable}%
    }
\caption{TurkCorpus}
\end{subtable}%

\bigskip

\begin{subtable}{.48\textwidth}
    \centering
    \resizebox{\textwidth}{!}{%
    \begin{tabular}{l|c|ccc|c}
        \toprule
            & SARI$\uparrow$ & ADD$\uparrow$ & DELETE$\uparrow$ & KEEP$\uparrow$ & FKGL$\downarrow$  \\
        \midrule
        \midrule
        \textbf{Recent Works} \\
        \midrule
        \midrule
        \citet{martin-etal-2020-controllable} & 40.13	& 	-	& -	& 	-	& 	7.29  \\
        \citet{martin2020multilingual} & 44.15$_{\pm 0.6}$ & - & - & - & 7.60$_{\pm 1.06}$  \\
        \midrule
        \midrule
        Reference Baseline & \textbf{44.89$_{\pm 0.90}$} & \textbf{10.17$_{\pm 1.20}$} & 58.76$_{\pm 2.24}$ & \textbf{65.73$_{\pm 2.03}$} & \textbf{6.49$_{\pm 0.42}$} \\
        \midrule
        \midrule
        \textbf{Our System} \\
        \midrule
        \textsc{TST-Base} & 37.4$_{\pm 1.62}$ & 3.62$_{\pm 0.59}$ & 47.22$_{\pm 4.5}$ & 61.37$_{\pm 0.52}$ & 8.08$_{\pm 0.31}$ \\
        \textsc{TST-Final}  & 43.21$_{\pm 0.3}$ & 8.04$_{\pm 0.29}$ & \textbf{64.25$_{\pm 1.22}$} & 57.35$_{\pm 1.68}$ & 6.87$_{\pm 0.27}$  \\
        \bottomrule
    \end{tabular}
    }
    \caption{ASSET}
\end{subtable}%

\bigskip

\begin{subtable}{.48\textwidth}
    \centering
    \resizebox{\textwidth}{!}{%
    \begin{threeparttable}[t]
    \begin{tabular}{l|c|ccc|c}
        \toprule
            & SARI$\uparrow$ & ADD$\uparrow$ & DELETE$\uparrow$ & KEEP$\uparrow$ & FKGL$\downarrow$  \\
    \midrule
    \midrule
    \textbf{Recent Works} \\
    \midrule
    \midrule
    \citet{zhang-lapata-2017-sentence} & 27.24 & - & - & - & 7.55 \\
    \citet{alva-manchego-etal-2017-learning}\tnote{$\ddagger$} & 30.50 & 2.72 & 76.31 & 12.46 & 9.38  \\
    \citet{vu-etal-2018-sentence} &  29.75 & - & - & - & - \\
    \citet{guo-etal-2018-dynamic} & 28.24 & - & - & - & 6.93 \\
    \citet{qiang2018improving} & 26.49 & - & - & - & 10.75 \\
    \citet{dong-etal-2019-editnts} & 32.35 & 2.24 & \textbf{81.30} &13.54  & \textbf{5.47} \\
    \citet{zhao2020semi} & 	36.92	& 	2.04	& 	72.79	& 	35.93	& 	- \\
    \midrule
    \midrule
    Reference Baseline & - & - & - & - & 8.74 \\
    \midrule
    \midrule
    \textbf{Our System} \\
    \midrule
    \textsc{TST-Base} & 43.11$_{\pm 1.87}$ & 4.66$_{\pm 1.31}$ & 61.13$_{\pm 4.73}$ & \textbf{63.54$_{\pm 2.75}$} & 8.41$_{\pm 1.01}$ \\
    \textsc{TST-Final} & \textbf{44.67$_{\pm 1.26}$} & \textbf{8.12$_{\pm 0.92}$} & 64.87$_{\pm 2.09}$	& 61.01$_{\pm 1.76}$ & 9.29$_{\pm 0.9}$ \\
    \bottomrule
    \end{tabular}
    \begin{tablenotes}
        \item[$\ddagger$] Quoted from the re-implementation by \citet{dong-etal-2019-editnts}.
    \end{tablenotes}
    \end{threeparttable}%
    }
\caption{WikiSmall}
\end{subtable}%

\caption{Comparison of our system against recent state-of-the-art Neural Text Simplification models on TurkCorpus, ASSET and WikiSmall test sets.}

\label{table:results}
\end{table}

\subsection{Text Simplification and Readability}

Table \ref{table:results} summarizes the results of our evaluations on TurkCorpus, ASSET and WikiSmall test sets. To ensure robustness of results, we report average scores of 4 runs with different random seeds. We compare the results of our baseline model (\textsc{TST-Base}) and our final model (\textsc{TST-Final}) against recent state-of-the-art Neural  Text  Simplification models. Additionally, we compare against a reference baseline similar to  \citet{martin2020multilingual}, where we 
compute the scores in a leave-one-out scenario where each reference is evaluated against all the others and then scores are averaged over all references. \textsc{TST-Final} consists of all the enhancements mentioned in Section \ref{sec:experimental_setup} added on top of \textsc{TST-Base}: data pre-processing, GEC-initialization, data augmentation, and inference tweaks.
In terms of the FKGL score, our system achieves better results than the reference baselines on TurkCorpus, and comes within 0.5 points on ASSET and WikiSmall. Compared to the state-of-the-art \cite{martin2020multilingual}, it improves by 0.23 FKGL points on average, indicating that the simplifying edits made by TST are easier to understand. 

In terms of SARI metrics, 
\textsc{TST-Base} achieves a competitive score of 39.17 on TurkCorpus, and a state-of-the-art SARI score of 43.11 on WikiSmall, outperforming the previous state-of-the-art result by a huge margin of 6.19 SARI points. This shows that simply using our baseline architecture to train a Text Simplification model on WikiAll can achieve competitive performance on the task. On the other hand, the \textsc{TST-Final} makes significant improvements over \textsc{TST-Base}. On TurkCorpus and ASSET, it comes within 1 SARI point of the current state-of-the-art \cite{martin2020multilingual}, outperforming all other prior text simplification models in literature. On WikiSmall, it further improves its state-of-the-art performance from \textsc{TST-Base} to achieve a SARI score of 44.67. 

It can be seen that compared to prior works, the most significant improvements in both the  \textsc{TST} models come from \textsc{add} and \textsc{delete} operations. It is noteworthy that \textsc{TST-Final} is able to achieve the highest F1 scores on \textsc{add} (6.96) and \textsc{delete} (47.87) SARI operations reported in literature on TurkCorpus. On the ASSET dataset, the F1 scores on \textsc{add} and \textsc{delete} operations improve further to 8.04 and 64.25 respectively, improving by large margins over \textsc{TST-Base}.
Similarly, it outperforms the state-of-the-art on \textsc{add} operations on WikiSmall. This shows that models proposed in prior works learned a safe, but inefficient strategy of simplification - leaning heavily on copying the sources sentences directly, owing to their high \textsc{keep} scores. By contrast, our model learns to edit the sentences better, as shown by the lower rates of keeping the source sentences unchanged. This is further verified by the fact that outputs of prior works\footnote{Measured on TurkCorpus. This information was not available for ASSET and WikiSmall} have much longer output sentence lengths (avg. 19.26 words) compared to ours (avg. 16.7 words), leading to more effective simplifications.





\subsection{Inference Time}

We also compare our system’s inference times against the current state-of-the-art text simplification systems. Specifically, we compare against ACCESS (trained on WikiLarge data) \cite{martin-etal-2020-controllable} and \textsc{BART+ACCESS} \cite{martin2020multilingual} (trained on WikiLarge + \textsc{mined} data) systems. We used the publicly available model checkpoint for ACCESS to compare against \citet{martin-etal-2020-controllable}. 
Direct comparison against \textsc{BART+ACCESS} was not possible because of the lack of publicly available code. Therefore, we used \textsc{BART} \cite{lewis-etal-2020-bart} for text summarization as a proxy for \citet{martin2020multilingual}.
We ran all systems with batch size 128 on the TurkCorpus test set 100 times, using NVIDIA Tesla V100. Within a single run, the results were averaged across all batches. We took into account only the actual inference time and omitted any initialization times. 

\begin{table}
\begin{center}
\small
\begin{tabular}{lc}
\toprule
\textbf{System} & \textbf{Inference time (sec)} \\
\toprule
BART, beam size = 8 & 2.82\\
BART, beam size = 2 & 1.95\\
ACCESS, beam size = 8 & 1.43\\
ACCESS, beam size = 1 & 1.14\\
\midrule
TST, 5 iterations & 0.43\\
TST, 4 iterations & 0.39\\
TST, 3 iterations & 0.33\\
TST, 2 iterations & 0.24\\
TST, 1 iteration & 0.13\\
\bottomrule
\end{tabular}
\caption{Average inference time per batch. In the context of TST, iterations refers to the number of iterations mentioned in Section \ref{ssec:iterative_tagging}}
\label{tab:inference-time}
\end{center}
\end{table}

The results in Table~\ref{tab:inference-time} show that the inference speeds\footnote{We compared ACCESS and BART with beam size 8 and TST with 2 iterations, as reported on TurkCorpus} of TST are at least 6 times faster than \textsc{ACCESS} and 11.75 times faster than pure \textsc{BART} which is the crucial component of the current state-of-the-art \cite{martin2020multilingual}. The impact of a non-autoregressive model architecture can be clearly seen here since TST is a sequence tagging system and does not need to predict edits one-by-one as done by auto-regressive transformer decoders (like the one used in \textsc{ACCESS}). Therefore, the inference is naturally parallelizable and therefore runs many times faster.

\section{Ablation Study}

\label{sec:ablation_studies}
In this section, we present ablation experiments for each of the enhancements described in Section \ref{sec:experimental_setup}, and applied to \textsc{TST-Base} to obtain \textsc{TST-Final}. The results of these experiments (Table \ref{tab:sari-ablation}) are reported on SARI and FKGL scores, averaged between ASSET and TurkCorpus test datasets. Each result is reported using an average of 4 runs for each experiment. Overall, the enhancements improve the SARI score by 4.0 points and FKGL by 0.21 points, while reducing variance in both cases. 

\subsection{GEC Initialization}
\label{ssec:ablation_gec}
We improve our strong baseline model \textsc{TST-Base} by pre-training it on the GEC task.\footnote{We refer the reader to \citet{omelianchuk-etal-2020-gector} for details on training the model for GEC.} Even though we find that using \textsc{TST-GEC}
leads to modest immediate improvements (+0.1 SARI point) compared to \textsc{TST-Base}, we found that adding other enhancements without the GEC-pre-training were not as effective, with the final model (\textsc{TST-Base} + Filtering + WikiEns + InfTweaks) achieving an average SARI score of 40.01 - significantly lower than the one with GEC-pre-training (42.3). 

These results show that pre-training \textsc{TST-Base} on GEC is an effective way to initialize the model for Text Simplification, since it equips the model to make additions and deletions, which are then further improved during training on the text simplification data. This was also not unexpected because the edit-tags were obtained from huge amounts of GEC data, and are expected to have good coverage with regards to many standard monolingual text editing problems - as also observed by a high overlap in the tag distributions between the GEC and Text Simplification domains (Section \ref{section:transformations}). 

\subsection{Data Pre-processing}
\label{ssec:ablation_brackets}

\begin{table}
    \begin{center}
    \small
    \begin{tabular}{lcc}
        \toprule
        \textbf{System} & \textbf{SARI} $\uparrow$ & \textbf{FKGL} $\downarrow$ \\
        \toprule
        TST & 38.3  $\pm$ 1.36 & 8.08 $\pm$ 0.31 \\
        \hspace{5pt} + GEC  & 38.4 $\pm$ 0.83 & 8.32 $\pm$ 0.26 \\
        \hspace{5pt} + Filtering  & 39.1 $\pm$ 0.48 & 7.66 $\pm$ 0.25 \\
        \hspace{5pt} + WikiBT & 39.5 $\pm$ 0.01 & 7.5 $\pm$ 0.06 \\
        \hspace{5pt} + WikiEns (- WikiBT) & 40.3 $\pm$ 0.15 & \textbf{7.48 $\pm$ 0.2} \\
        \hspace{5pt} + InfTweaks & \textbf{42.3 $\pm$ 0.25} & 7.87 $\pm$ 0.19 \\
        \bottomrule
    \end{tabular}
    \caption{Average SARI and FKGL scores (ASSET and TurkCorpus test sets)}
    \label{tab:sari-ablation}
\end{center}
\end{table}

As mentioned in Section \ref{ssec:data_preproc}, we removed special tokens found in Wikipedia data such as \texttt{-LRB-} and \texttt{-RRB-}, along with the text enclosed by these tokens, in both source and target sentences. 
We find that 
using GEC initialization together with filtering brackets was beneficial to the system (+0.8 SARI points), and also decreased the variance in the results. The benefit of this step is towards improving text simplification quality is also seconded by a significantly reduced FKGL score (-0.66 points).

\subsection{Data Augmentation}
We explored two strategies of data augmentation: enriching training data with (i) back-translated data (WikiBT), (ii) ensemble-generated data (WikiEns). Augmenting the training data using WikiEns leads to a bigger boost compared to just adding WikiBT (+1.2 vs +0.4). We also experimented with adding both synthetic datasets (WikiEnd + WikiBT) to the WikiAll training data, but the performance was worse compared to using only WikiEns. 

\subsection{Inference Tweaks}
Finally, we describe the effect of tuning the inference hyper-parameters for our model obtained so far. Using these tweaks (Section \ref{ssec:inference_tweaks_description}) is one of the most crucial components of our system. Overall, it is not just able to affect the sequence generation, but also gives us the biggest boost (+2.0 points). Our final model with inference tweaks comfortably outperforms its predecessor on all datasets, demonstrating their effectiveness on the task.

\section{Conclusion}
This paper introduces TST, a novel approach to text simplification, by reformulating the task into a much simpler one of sequence tagging. We build TST by adapting the GECToR framework for GEC. We show that most of its performance gains are owed to simplistic data augmentations and tweaks in training and inference. These modifications allow us to derive maximal benefit from the already existing pre-trained Transformer-based encoders on large amounts of human-annotated and synthetic data, making TST a simple, powerful, easily reusable method for monolingual editing tasks. Since TST is able to progressively make simplifying edits via explicit edit-tag operations, the transformations resulting from TST are better explainable and interpretable than any other NMT-based Seq2Seq approaches. Finally, TST is fully non-autoregressive, enabling it to perform faster inference than any other state-of-the-art text simplification methods. Our empirical results demonstrate that it achieves near state-of-the-art performance on benchmark test datasets for text simplification.

A major motivation in this work was to minimize changes to the original model to keep the system simple, fast, and reproducible. Hence, we restricted our system to only use WikiAll data and its derivatives (vs. any external data like the state-of-the-art system by  \citet{martin2020multilingual}). While we did not fully beat the state-of-the-art on the task, we believe that using larger models (eg. RoBERTa$_{\textsc{LARGE}}$), ensembles, or external data will likely lead to better SARI scores at the cost of speed and system complexity: ideas we plan to explore in future work. 


\bibliography{anthology,eacl2021}
\bibliographystyle{acl_natbib}

\clearpage

\appendix

\section{Model Configurations}
\label{sec:appendix1}
Table \ref{tab:hyperparameters} describes the list of hyper-parameters used for \textsc{TST-Final} model. In Table \ref{tab:inference_tweaks}, we list the inference tweaks hyper-parameters found by Bayesian Search on TurkCorpus and ASSET datasets. 


\begin{table}[H]
\begin{center}
\small
\begin{tabular}{lc}
\toprule
\textbf{Hyperparameter name} & \textbf{Value} \\
\toprule
learning\_rate & 1e-5\\
transformer\_model & roberta-base \\
accumulation\_size & 2 \\
batch\_size & 128 \\
cold\_steps\_count & 2 \\
n\_epoch & 50 \\
patience & 3 \\
vocab\_size & 5000 \\
max\_len & 50 \\
min\_len & 3 \\
pieces\_per\_token & 5 \\
filter\_brackets & 0/1 \\
seed & 1/2/3/11 \\
normalize & 1 \\
\bottomrule
\end{tabular}
\caption{The list of hyper-parameters used during training}
\label{tab:hyperparameters}
\end{center}
\end{table}

\begin{table*}[b]
\begin{center}
\small
\begin{tabular}{ccccccc}
\toprule
\textbf{Model description} & \textbf{Tuned dataset} & \textbf{Seed} & \textbf{Del conf} &	\textbf{Keep conf} & \textbf{Iterations} &	\textbf{Min error probability}  \\
\toprule
TST w/o InfTweaks & - & - & 0 & 0 & 5 & 0 \\
TST with InfTweaks & Turk & 1 & -0.84 &	-0.66 &	2 & 0.04 \\
TST with InfTweaks & Turk & 2 &	-0.93 &	-0.51 &	3 &	0.02 \\
TST with InfTweaks & Turk & 3	&-0.86 &	-0.68 &	2 &	0.03 \\
TST with InfTweaks & Turk & 11	& -0.88 & -0.61 &	2 &	0.03 \\
TST with InfTweaks & ASSET & 1 & -0.66 &	-0.9 &	3 & 0.02 \\
TST with InfTweaks & ASSET & 2 & -0.72 &	-0.88 &	3 &	0.02 \\
TST with InfTweaks & ASSET & 3	& -0.52 &	-0.89 &	3 & 0.04 \\
TST with InfTweaks & ASSET & 11	& -0.72 &	-0.91 &	3 &	0.02 \\

\bottomrule
\end{tabular}
\caption{Inference hyper-parameters found by the Bayesian Search on TurkCorpus and ASSET development sets}
\label{tab:inference_tweaks}
\end{center}
\end{table*}

\newpage

\section{Simplification Examples}

Various examples from our system are shown in Table \ref{tab:examples}. Examining the simplifications, we see reduced sentence length, sentence splitting of a complex sentence into multiple shorter sentences, and the use of simpler vocabulary. Manual comparison between \textsc{TST-base} and \textsc{TST-Final} shows that the first system tends to delete some complex words from the text. For example, ``theoretically possible” gets shortened to just ``possible,” and ``administrative district” to ``district”. \textsc{TST-Final} model tends to be more creative and changes phrases to simpler versions like “is theoretically possible” to ``might be” or ``an administrative district” to ``a part of.” However, this aggressive and creative strategy sometimes also generates ungrammatical output like in the last example in Table \ref{tab:examples}. While it rarely happens, but the model might also change the meaning of the original sentence. For example, replacing ``the five” to ``the three.” It is worth noticing that the same problem of meaning change is present in the reference sentences as well: where ``the five” got replaced with ``one of four”.


\begin{table*}[t]
\centering
\resizebox{\textwidth}{!}{%
    \begin{tabular}{p{1.9cm}|p{18cm}}
        \toprule
        Original & he also \textbf{completed two collections} of short stories \textbf{entitled} the ribbajack \& other curious yarns and seven strange and ghostly tales . \\
        Reference & he also \textbf{wrote two books} of short stories \textbf{called ,} the ribbajack \& other curious yarns and seven strange and ghostly tales . \\
        \textsc{TST-Base} & he also  \textbf{wrote} two collections of short stories  \textbf{called} the ribbajack \& other curious yarns and seven strange and ghostly tales . \\
        \textsc{TST-Final} & he also \textbf{wrote a series} of short stories \textbf{called} the ribbajack \& other curious yarns and seven strange and ghostly tales . \\
        \midrule
        Original & it \textbf{is theoretically possible} that the other editors who may have \textbf{reported} you , and the administrator who blocked you , are part of a conspiracy against someone half a world away they 've never met in person . \\
        Reference & it is theoretically possible that the other editors who may \textbf{have written about} you, and the \textbf{officer} who blocked you, are part of \textbf{a bad plan} against someone \textbf{miles} away, they 've never met \textbf{face to face} . \\
        \textsc{TST-Base} & it is possible that the other editors who may have reported you , and the administrator who blocked you , are part of a conspiracy against someone half a world away they ' ve never met in person . \\
        \textsc{TST-Final} & it \textbf{might be} that the other editors who may have \textbf{sent} you , and the administrator who blocked you , are part of a conspiracy against someone half a world away where they 've never met in person . \\
        \midrule
        Original & as a result , although many mosques will not enforce violations , both men and women when attending a mosque must \textbf{adhere to these guidelines} . \\
        Reference & as a result , both men and women must follow this rule when they attend a mosque , even though many mosques do not enforce these rules  \\
        \textsc{TST-Base} & both men and women when \textbf{going} a mosque must \textbf{follow these rules} . \\
        \textsc{TST-Final} & both men \textbf{,} and women \textbf{that attend} mosque \textbf{,} must \textbf{follow the law} . \\
        \midrule
        Original & hinterrhein is \textbf{an administrative district in} the canton of graubünden , switzerland . \\
        Reference & hinterrhein is a district \textbf{of} the canton of graubünden , switzerland . \\
        \textsc{TST-Base} & hinterrhein is a district \textbf{of} the canton of graubünden , switzerland . \\
        \textsc{TST-Final} & hinterrhein is \textbf{a part of} the canton of graubünden in the switzerland . \\
        \midrule        
        Original & a majority of south indians speak one of the five dravidian languages — kannada , malayalam , tamil , telugu and tulu . \\
        Reference & \textbf{many} of the south indians \textbf{are dravidians and they speak one of four} dravidian languages — kannada , malayalam , tamil \textbf{or} telugu . \\
        \textsc{TST-Base} & \textbf{most} of south indians speak one of the five dravidian languages — kannada , malayalam , tamil , telugu and tulu . \\
        \textsc{TST-Final} & \textbf{most of the people} speak \textbf{speakers from the three} dravidian languages \textbf{spoken are ,} kannada , malayalam , tamil , telugu \textbf{,} and tulu . \\
        
        \bottomrule
    \end{tabular}
}
\caption{Examples of simplifications by TST}
\label{tab:examples}
\end{table*}

\end{document}